# Modeling of Human Criminal Behavior using Probabilistic Networks

Ramesh Kumar  Gopala Pillai
Research Scholar
R.V. Center for Cognitive Technologies
Bangalore, India
-

Dr. Ramakanth Kumar .P
Professor
R.V. Center for Cognitive Technologies
Bangalore, India
-

*Abstract— Currently, criminal's profile (CP) is obtained from investigator's or forensic psychologist's interpretation, linking crime scene characteristics and an offender's behavior to his or her characteristics and psychological profile. This paper seeks an efficient and systematic discovery of non-obvious and valuable patterns between variables from a large database of solved cases via a probabilistic network (PN) modeling approach. The PN structure can be used to extract behavioral patterns and to gain insight into what factors influence these behaviors. Thus, when a new case is being investigated and the profile variables are unknown because the offender has yet to be identified, the observed crime scene variables are used to infer the unknown variables based on their connections in the structure and the corresponding numerical (probabilistic) weights. The objective is to produce a more systematic and empirical approach to profiling, and to use the resulting PN model as a decision tool.*

## I.  INTRODUCTION

Modeling human criminal behavior is challenging due to many variables involved and the high degree of uncertainty surrounding a criminal act and the corresponding investigation. Probabilistic graphs are suitable modeling techniques because they are inherently distributed and stochastic. In this paper, the system variables comprising the PN are offender behaviors and crime scene evidence, which are initialized by experts through their professional experience or expert knowledge.

The mathematical relationships naturally embedded in a set of crimes [3, 4, 8] are learned through training from a database containing solved criminal cases. The PN model is to be applied when only the crime scene evidence is known to obtain a useable offender profile to aid law enforcement in the investigations. A criminal profile is predicted with a certain quantitative confidence.

The PN approach presented here seeks to build on the ideas of behavior correlations in order to obtain a usable criminal profile when only crime scene evidence is known from the investigation.

This paper proposes a systematic approach for deriving a multidisciplinary behavioral model of criminal behavior. The proposed offender behavioral model is a mathematical representation of a system comprised of an offender's actions and decisions at a crime scene and the offender's personal characteristics.

The influence of the offender traits and characteristics on the resulting crime scene behaviors is captured by a probabilistic graph or PN that maps cause-and-effect relationships between events, and lends itself to inductive logic for reasoning under uncertainty [1]. The use of PNs for CP may allow investigators to take into consideration various aspects of the crime and discover behavioral patterns that might otherwise remain hidden in the data. The various aspects of a crime include a victimology assessment (victim's characteristics, e.g., background characteristics, age, gender, and education), crime scene analysis (evidence from the crime scene, e.g., time and place where the crime occurred), and a medical report (autopsy report, e.g., type of non-deadly and deadly lesions and signs of self defense).

The PN approach to criminal profiling is demonstrated by learning from a series of crime scene and offender behaviors. The learning techniques employed in this modeling research are evaluated on a set of validation cases not used for training by defining a prediction accuracy based on the most likely value of the output variables (offender profile) and its corresponding confidence level.

## II.  APPROACH

To start with, a graphical model of offender behavior is learned from a database of solved cases. The resulting CP model obtained through training is then tested by comparing its predictions to the actual offenders' profiles.

Let the database sample space = D,

Let D consist of 'd' solved cases $\{C_1, ..., C_d\}$,

where $C_i$ is an instantiation of X, which is randomly partitioned into two independent datasets such as a training set

T and a validation set V, such that D = T $U$ V. The variables in X are partitioned as follows: the inputs are the crime scene (CS) variables $X^I$ (evidence) for $X^I = (X^I_1, ..., X^I_k)$, and the outputs are the offender (OFF) variables comprising the criminal profile $X^O$, for $X^O = (X^O_1, ..., X^O_m)$, where $(X^I, X^O)$ ε X.

The PN model is learned from T, as explained later, and it is tested by performing inference to predict the offender variables (OFF) in the validation cases V. An offender profile is estimated based on crime scene evidence, with a prediction being the most likely value of a particular offender variable. During the testing phase, the predicted value of $X^O_i$, denoted by $x^P_{i,a}$ where a=1 or 2 for a binary variable, is compared to the observed state $x^O_{i,b}$ obtained from the validation set V, where b=1 or 2. An example of an offender variable is "gender", with states "male" and "female". The overall performance of the PN model is evaluated by comparing the true (observed) states $x^O_{i,b}$ to the predicted output variable values $x^P_{i,a}$ in the validation cases. This process tests the generalization properties of the model by evaluating its efficiency over V.

## III. VARIABLES CONSIDERED

The relevant categories of variables that have emerged from the criminal profiling research as selected by investigators, criminologists, and forensic psychologists are described as follows:

• *Crime Scene Analysis* (CSA): CSA variables are systematic observations made at the crime scene by the investigator. Examples of CSA variable pertain to where the body was found (e.g., neighborhood, location, environment characteristics), how the victim was found (e.g., the body was well-hidden, partially hidden, or intentionally placed for discovery), and the correlation between where the crime took place and where the body was found (e.g., the body was transported after the murder).

• *Victimology Analysis* (VA): VA variables consist of the background characteristics of the victim independent of the crime. For example, VA variables include the age, sex, race, education level, and occupation of the victim.

• *Forensic Analysis* (FA): FA variables rely on the medical examiner's report that deals with the autopsy. Examples of this are time of death, cause of death, type of non-lethal wounding, wound localization, and type of weapon that administered the wounds.

The set of CP variables used in this paper were defined in previous research [4, 5, 7, 8]. The selection criteria for variable selection [6] are:

**(i)** Behaviors are clearly observable and not easily misinterpreted

**(ii)** Behaviors are reflected in the crime scene, e.g., type of wounding, and

**(iii)** Behaviors indicate how the offender acted toward and interacted with the victim e.g., victim was bound/gagged, or tortured. Some crime scene (CS) variables describing the observable crime scene and offender (OFF) variables describing the actual offender were selected based on the above criteria. Examples of the CS variables are multiple wounding to one area, drugging the victim, and sexual assault. Examples of the offender variables include prior offenses, relationship to the victim, prior arrests, etc. The variables all have binary values representing whether the event was present or absent.

## IV. TRAINING THE MODEL

The basic schematic of the training software, including the validation process, is shown in Figure 1, where $P^h$ is the proposed PN and $P^{opt}$ is the trained (or optimized) PN. The software is intended to aid law enforcement in the investigation of violent crimes. Because the cases are unsolved and only the crime scene inputs are known, the criminal profiling software consists of a trained PN model that has been previously trained and validated with D. Also, the model has the potential to be updated by means of an incremental training algorithm when additional cases are solved by the police. Thus, $P^{trained}$ consistently reflects the model of an evolving criminal profile over time.

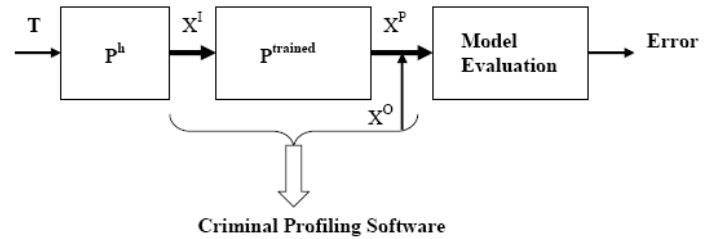

**Figure 1:** CP model training and validation software

## V. DATA BASE OF SOLVED CASES

A set of single offender/single victim homicides was collected by psychologists from solved homicide files of the British police [4, 6]. In order to examine the aggressive behavioral patterns of a particularly violent offense, the criteria for case selection is: single offender/single victim homicide cases; a mixture of domestic (where the victim and offender were known to each other (e.g., family member, spouse, co-worker) and stranger (the offender is unknown to the victim, thus they had no previous links to each other) cases; offenders are adults at least 17 years of age, as defined by the court system. Excluded from the sample were cases when the cause of death was not aggressive or extremely intentional. Homicides by reckless driving are not included due to the lack of interpersonal interaction between the victim and offender.

## VI. SAMPLING

A simulation set is built to produce an artificial CP database to study the PN learning and inference capabilities. This included a more extensive list of crime scene, offender characteristics and multiple-valued variables. A PN is used to simulate a set of cases where the crime scene and offender variables can be chosen by the user. An initial structure thus relating the variables and the corresponding initial probabilistic parameters $\theta_0$ are declared based on the prior knowledge, through experience, or by sampled statistics. Cases are simulated by feed forward sampling, where variables are sampled one at a time in order from top-level variables (variables without parents), to the mid-level variables (variables with both parents and children), ending with the bottom-level variables (children variables with parents only). For each variable, the discrete conditional prior probabilities in vector form are given as:

$$[P(x_{i,1}|\ \pi_i),\ (x_{i,2}|\ \pi_i),\ \ldots\ldots,\ (x_{i,ri}|\ \pi_i)]$$

where $r_i$ is the maximum state for $X_i$ and $\pi_i$ disappears if $X_i$ is a top-level variable. A value $v_i$ is drawn from a uniform continuous distribution between '0' and '1' and the conditional prior probability vector as a vector of ranges becomes

$$[P(x_{i,1}|\ \pi_i),P(x_{i,1}|\ \pi_i)+\ P(x_{i,2}|\ \pi_i),\ \ldots\ldots,$$

$$\sum_{i=1}^{r_i} (x_{i,j}|\ \pi_i)]\ \ which\ refers\ to$$

$$X_i = \begin{cases} x_{i,1} \text{ if } 0 < v < P(x_{i,1}|\pi_i) \\[2mm] x_{i,2} \text{ if } P(x_{i,1}|\pi_i) \leq v < \sum_{i=1}^{r_i} P(x_{i,j}|\pi_i) \\[2mm] x_{i,ri} \text{ if } \sum_{i=1}^{r_i} P(x_{i,j}|\pi_i) \leq v < 1 \end{cases} \tag{1}$$

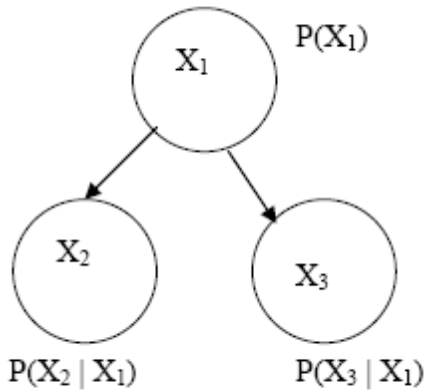

**Figure 2: Three-nodal model**

To simulate a set of cases for the system represented by the three-nodal model illustrated in Figure 2, the variables are ordered as $(X_1, X_2, X_3)$, where $X_1$ is the parent of $X_2$ and $X_3$, and $X_1 = (x_{1,1}, x_{1,2})$, $X2 = (x_{2,1}, x_{2,2})$ and $X3 = (x_{3,1}, x_{3,2})$. Starting with $X_1$, it has three possible states with the prior probabilities $P(x_{1,1}) = 0.2$, $P(x_{1,2}) = 0.5$, and $P(x_{1,3}) = 0.3$, which becomes a vector [0.2, 0.7, 1] referring to:

$$X_1 = \begin{cases} x_{1,1} \text{ if } 0 \leq v_1 < 0.2 \\[2mm] x_{1,2} \text{ if } 0.2 \leq v_1 < 0.9 \\[2mm] x_{1,3} \text{ if } 0.9 \leq v_1 < 1 \end{cases} \tag{2}$$

If $v_1 = 0.11$ which makes $X_1 = x_{1,1}$ and the Conditional Probability Table (CPT) for X2 is listed in Table 1.

| $X_2$ | $P(x_{2,1}|\ X_1)$ | $P(x_{2,2}|\ X_1)$ |
|---|---|---|
| $X_1 = x_{1,1}$ | 0.2 | 0.8 |
| $X_1 = x_{1,2}$ | 0.9 | 0.1 |

**Table 1 Conditional probability Table**

Then the conditional prior probability vector of ranges for a newly generated $v_2$ becomes

$$X_2 = \begin{cases} x_{2,1} \text{ if } 0 \leq v_2 < 0.2 \\[2mm] x_{2,2} \text{ if } 0.2 \leq v_2 < 1 \end{cases} \tag{3}$$

$X_3$ is sampled following the same procedure as $X_1$ and $X_2$. This is repeated until the desired number of cases as specified by the user is reached. The Matlab function utilized for the sampling exercise is sample b_net in the Bayes Net Toolbox [2].

## VII. PN PREDICTIONS AND ACCURACY

When a PN model of offender behavior on the crime scene is learned from solved cases, it is implemented on a set of solved validation cases in order to test the trained model's performance. Performance is tested through probabilistic inference. Inference is the process of updating the probability distribution of a set of possible outcomes based upon the relationships represented by the PN model and the observations of one or more variables. With the updated probabilities, a prediction can be made from the most likely value of each inferred variable. Thus, in order to test the trained model, only the crime scene evidence is inserted into the model, with the predicted offender profile being compared to the actual offender characteristics. Because this is a probabilistic model, a certain confidence accompanies the offender variable predictions.

## VIII. CONCLUSIONS

This paper presents an approach for deriving a network model of criminal profiling that draws on knowledge-based systems and on fields of criminology and offender profiling. Implementing probabilistic networks makes it possible to represent multidimensional interdependencies between all relevant variables that have been identified in previous research as playing a role in determining or reflecting the behavior of offenders at the crime scene. Hence, a valid network model can be used to predict unknown variables composing an offender profile based on the variables observed from the crime scene.

characteristics from crime scene behavior. Scandinavian Journal of Psychology, 44:107–118, 2003.

## AUTHORS PROFILE

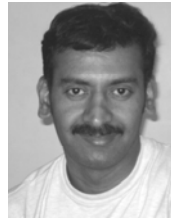

**Mr. Ramesh Kumar Gopala Pillai** is a post graduate in Information Technology from Kuvempu University. He also possesses a MBA degree in Banking and Finance. Currently he is pursuing his research in the area 'Modeling of Human Criminal Behavior and its Implications' under Kuvempu University. He has published a couple of papers in reputed International conferences and journals.

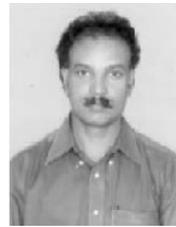

**Dr. Ramakanth Kumar P.** is a doctorate from Mangalore University. He has successfully guided research scholars to their PhD degree. He has executed several research projects as Principal Investigator for Government and Research establishments like DRDO, ISRO etc. He has several publications in reputed refereed International Journals and Conferences. Currently he is working as a Professor at R V Center for Cognitive Technologies, Bangalore.